\documentclass[11pt]{article}

\usepackage[preprint]{acl}
\definecolor{ctxgray}{gray}{0.95}
\definecolor{diffneg}{RGB}{180, 50, 50} 
\definecolor{diffpos}{RGB}{50, 150, 50} 
\definecolor{arrowred}{RGB}{180, 50, 50}
\definecolor{arrowgreen}{RGB}{50, 150, 50}
\definecolor{ctxblue}{RGB}{245, 245, 255}
\newcommand{\nodiff}{\multicolumn{1}{c}{--}}

\newcommand{\diffdown}[1]{\textcolor{arrowred}{$\downarrow$ #1}}
\newcommand{\diffup}[1]{\textcolor{arrowgreen}{$\uparrow$ #1}}

\usepackage{hyperref}
\usepackage{times}
\usepackage{latexsym}
\usepackage{booktabs}   
\usepackage{multirow}   
\usepackage[table]{xcolor} 
\usepackage[T1]{fontenc}

\usepackage[utf8]{inputenc}

\usepackage{microtype}
\usepackage{amsmath}
\usepackage{tcolorbox}
\tcbuselibrary{skins,breakable}

\usepackage{inconsolata}
\usepackage{subcaption}

\usepackage{graphicx}

\title{Presupposition and Reasoning in Conditionals:\\[0.3em] A Theory-Based Study of Humans and LLMs}

\author{
  \textbf{Tara Azin}\textsuperscript{1}\thanks{Corresponding author: \href{mailto:taraazin@cmail.carleton.ca}{taraazin@cmail.carleton.ca}},
  \textbf{Yongan Yu}\textsuperscript{2,3},
  \textbf{Raj Singh}\textsuperscript{1},
  \textbf{Olessia Jouravlev}\textsuperscript{1}
\\
\\
  \textsuperscript{1}Department of Cognitive Science, Carleton University \\
  \textsuperscript{2}School of Computer Science, McGill University\\
  \textsuperscript{3}Mila -- Quebec AI Institute\\
}

\begin{document}
\maketitle

\begin{abstract}

Presupposition projection in conditionals is central to theories of meaning and pragmatics, yet it remains largely unevaluated in large language models. We address this gap through a parallel behavioral study comparing human judgments and LLM predictions on a normed dataset of conditional sentences that controls the relation between the antecedent and the projected presupposition. We collect likelihood ratings from 120 participants and four LLMs under matched contextual conditions. Results show that humans integrate probabilistic and pragmatic cues in their judgment, whereas LLMs show variable alignment with human patterns. Using a linguistically motivated checklist within an LLM-as-a-Judge framework, we further evaluate model reasoning. We observe models that best match human ratings often lack coherent pragmatic reasoning, while models with stronger reasoning produce less human-like judgments. These findings suggest that LLMs' performance on such tasks may result from surface pattern matching rather than pragmatic competence. Our findings highlight the importance of benchmarks grounded in linguistic theory for comparing humans and models.

\end{abstract}

\section{Introduction}

Understanding how implicit meaning is inferred in context has remained a complicated challenge for theories of natural language interpretation. A classic puzzle is the proviso problem \citep{Geurts-1996} in presupposition projection, which concerns how presuppositions triggered in the consequent of conditionals are interpreted. In some cases, the presupposition is clearly understood as unconditional (e.g., \textit{If John flies to London, his sister will pick him up} presupposes that John has a sister). In other cases, however, the presupposition may be understood either unconditionally or conditional on the antecedent (e.g., \textit{If John is a scuba diver, he will bring his wetsuit}).
According to satisfaction theories of presupposition projection \citep{Heim-1983, beaver2001presupposition, schlenker2008presupposition}, a sentence of the form $\textit{If } A, B_p$, where $B$ contains a presupposition trigger, and $p$ is the presupposition of $B$, projects the conditional presupposition $A \rightarrow p$. In cases where accommodation is required\footnote{Accommodation refers to the pragmatic process by which listeners update the common ground to satisfy a presupposition when it is not already entailed.} \citep{Lewis-1979, vonFintel-2008}, listeners may either accommodate $A \rightarrow p$ or strengthen it to $p$ itself \citep{Singh2007, Singh-2020}. The proviso problem thus concerns how this accommodation decision is made and what role plausibility and context-sensitive reasoning play in determining whether $p$ is interpreted conditionally or unconditionally \citep{Mandelkern-2016, mandelkern2019independence}. Despite decades of theoretical work, no consensus has been reached on this problem.

Recent work evaluates the linguistic capabilities of large language models (LLMs) \citep{hale-stanojevic-2024-llms, sogaard-2025-language}, but existing studies suggest that they often struggle with semantic and pragmatic reasoning tasks compared to human judgments \citep{sravanthi-etal-2024-pub}. From a linguistic perspective, as noted above, presupposition reasoning reflects the interaction of semantics, pragmatics, discourse context, and probabilistic reasoning, and is sensitive to relevance and world knowledge \citep{Mandelkern-2016, Domaneschi2016}. However, these interactions have not yet been examined in controlled human--LLM comparisons.


In this paper, we present a comparison of human and LLM presupposition judgments based on the proviso problem in conditional sentences. We investigate how probabilistic relevance between the antecedent and the presupposition affects projection patterns in conditionals of the form $\textit{If } A, B_{\textit{p}}$, where $B$ contains a possessive trigger and $p$ is the presupposition of $B$. We construct a dataset of 90 sentences based on 30 base propositions, each instantiated in three variants that manipulate the logical and probabilistic relationship between $A$ and $p$. We collect likelihood ratings from 120 human participants and four LLMs on a 0--7 Likert scale (0 = very unlikely, 7 = very likely), with and without minimal contextual information. This design allows us to examine how presupposition judgments vary with $A$--$p$ relevance and how they are influenced by contextual information. The design is in line with current works using parallel human-LLM comparisons to assess linguistic competence \citep{qiu-etal-2024-evaluating}.

In addition, we conduct an LLM-as-a-judge experiment to analyze the reasoning underlying models' predictions. Using a theory-informed checklist that is grounded in formal semantics and pragmatic principles, a judge model evaluates whether explanations generated by the models reflect valid inferential patterns. Human experts are involved in both designing the checklist and in the meta evaluation stage. We argue that this approach moves beyond evaluation that is solely based on accuracy metrics toward a more interpretable analysis of model behavior, and is better suited to semantic and pragmatic tasks such as presupposition handling.

Our study addresses the following research questions: (i) How do human presupposition judgments in conditional sentences vary with antecedent-presupposition relevance? (ii) How closely do LLM judgments align with human patterns across contextual conditions? (iii) How does minimal discourse context influence interpretation for humans and models? (iv) To what extent do LLM explanations represent theoretically grounded reasoning about presupposition projection?\footnote{Codes and dataset are available at \href{https://github.com/proviso-bench/Presupposition-and-Reasoning-in-Conditionals}{https://github.com/proviso-bench/Presupposition-and-Reasoning-in-Conditionals}}

\section{Related Work}

\subsection{Human-Machine Presupposition Studies}

Research on presuppositional reasoning in language models is relatively recent. Prior work has examined conditional inference and presupposition judgment using prompt-based and NLI-style evaluations \citep{holliday-etal-2024-conditional, atwell-etal-2025-measuring}. Earlier studies evaluated models on datasets such as IMPPRES, NOPE, and PROPRES \citep{jeretic-etal-2020-natural, Parrish-2021, asami-sugawara-2023-propres}. These datasets mainly focus on entailment reasoning with simple conditional structures 
and do not address pragmatic factors that influence presupposition projection. As a result, they offer limited coverage of presupposition projection in complex conditionals. Overall, existing work primarily adopts classification-based evaluation and rarely conducts controlled behavioral comparisons with human judgments. Our study addresses this gap through a theory-driven human-LLM comparison of presupposition projection in conditionals.

\subsection{Pragmatic Evaluation of LLMs}

Recent work has examined whether LLMs exhibit pragmatic competence in areas such as implicature, presupposition, and reference \citep{jeretic-etal-2020-natural, kabbara-cheung-2022-investigating}. The Pragmatics Understanding Benchmark \citep{sravanthi-etal-2024-pub} provides the only large-scale theory-based resource with parallel human-LLM evaluation across multiple pragmatic phenomena. \citet{azin2025let} introduce CONFER, an NLI benchmark for this phenomenon, showing that models fail to generalize presuppositional reasoning to complex conditional structures. They further probe this behavior through explainability analyses, finding that models broadly align with human judgments on the proviso problem but rely on shallow pattern matching rather than pragmatic reasoning~\citep{azin2026language}. Our work builds on this line of research by focusing specifically on presupposition in conditional contexts and by providing a behavioral comparison based on pragmatic theories.

\subsection{LLM-as-a-Judge}

Traditional automatic metrics such as BLEU and ROUGE are limited in evaluating semantic and pragmatic reasoning \citep{papineni-etal-2002-bleu, lin-2004-rouge}. Recent work has therefore explored more interpretability-oriented LLM-as-a-Judge approaches \citep{zheng2023judging}.
\citet{lee-etal-2025-checkeval} propose a checklist-based framework that decomposes evaluation criteria into interpretable binary judgments. Building on this approach, we adopt and extend a theory-informed checklist tailored to presupposition inference.

\section{Methodology}

Our methodology consists of four stages: (1) a norming study to construct a controlled dataset, (2) collection of parallel presupposition judgments and reasoning from human participants and LLMs, (3) development of a checklist for evaluating pragmatic reasoning, and (4) automated evaluation using an LLM-as-a-Judge framework, followed by human validation.

\subsection{Data Construction and Norming}

Our dataset design is motivated by probabilistic accounts of presupposition accommodation in conditionals. Based on existing theories, when interpreting the presupposition ($p$) of a conditional sentence of the form $\textit{If } A, B_p$, where $p$ is the direct presupposition of the consequent $B$, listeners compare how likely $p$ is in general given background context $c$\footnote{By background context, we refer to shared world knowledge and situational assumptions available to interlocutors, such as general facts about people, places, and everyday situations.}, $Pr(p \mid c)$, with how likely $p$ is under the assumption that $A$ holds, $Pr(p \mid A, c)$~\cite{carnap1950logical, Strawson-1950, Stalnaker-1973, Stalnaker-1998}. In simple words, listeners consider whether assuming that $A$ holds makes $p$ much more likely to be true. If it does, they tend to interpret $p$ as holding only under the condition $A$; if it does not, they tend to interpret $p$ as generally true. For example, in \textit{If John is a scuba diver, then he will bring his wetsuit}, being a scuba diver ($A$) makes having a wetsuit ($p$) much more likely, whereas in \textit{If John flies to London, then his sister will pick him up}, flying to London ($A$) does not affect whether John has a sister ($p$). To operationalize this idea, we conduct a norming study to quantify and validate the probabilistic relevance between antecedents ($A$) and presuppositions ($p$). This allows us to obtain graded levels of $A$-$p$ relevance, which are used to construct the main study items for presupposition judgment.

We construct 30 base propositions corresponding to presupposed content (e.g., having a guitar, having an apron, having a boat). These propositions span high-probability ownerships (e.g., having a smartphone), moderate-probability cases (e.g., having siblings), and low-probability cases (e.g., having a boat). We also include neutral contextual constraints (e.g., someone being at a gym or at an airport) to restrict the possible world without directly affecting $p$.

For each proposition, we design four norming conditions: a baseline condition measuring $Pr(p \mid c)$ and three conditional criteria measuring $Pr(p \mid A, c)$ with high, mid, and low $A$–$p$ relevance antecedents. For example, for \textit{having a guitar} in the context \textit{someone at a gym}, we ask about the likelihood that the person (i) has a guitar (baseline), (ii) is a musician and has a guitar (high relevance), (iii) likes music and has a guitar (mid relevance), and (iv) speaks French and has a guitar (low relevance). Participants (thirty native English speakers residing in the US, Canada, and the UK, recruited via Prolific\footnote{\url{https://www.prolific.com}}) rated 120 items on a 1–7 Likert scale by judging how likely each statement was to be true, with items presented in randomized order. 

The norming results show clear statistical separation between low, mid, and high probability conditions, confirming our initial predictions. 
We therefore excluded the baseline items and retained the 90 conditional items for the main study. Each main study item followed the form $\textit{If } 
A, B_p$, with antecedents selected to induce relevant, somewhat relevant, or irrelevant $A$-$p$ relations, and with target propositions representing 
low-, mid-, and high-probability categories derived from the norming data (e.g., having a watch, having a smartphone).

For example, in \textit{If Daniel drinks tea, he will make tea for his sister}, the antecedent $A$ (\textit{drinks tea}) is semantically unrelated 
to the presupposition $p$ (\textit{Daniel has a sister}), yielding an irrelevant \emph{$A$-$p$} relation, and the proposition \textit{having a sister} 
corresponds to a mid-probability category based on norming results (see Appendix~\ref{app:stimuli} for more examples).

\begin{table*}[t]
\centering
\small
\renewcommand{\arraystretch}{1.2}
\begin{tabular}{lll}
\toprule
\textbf{Dimension} & \textbf{Theory Base} & \textbf{Examples of What It Tests} \\
\midrule

Accuracy 
& Discourse Representation Theory (DRT) 
& Trigger and anaphora detection  \\

Context 
&  Dynamic Semantics 
& Robustness to context effects\\

Pragmatic 
& Gricean Pragmatics 
& Relevance and common ground reasoning \\

Presupposition Handling 
& Projection and Accommodation Theory 
& Projection and accommodation control \\

Coherence 
& Argumentation Theory 
& Consistency between reasoning and judgments \\

\bottomrule
\end{tabular}

\caption{Checklist dimensions and representative evaluation criteria used by the LLM-as-a-judge model to assess LLM reasoning steps against semantic and pragmatic theories in conditional presupposition tasks.}

\label{tab:theoretical_framework}
\vspace{-3ex}
\end{table*}

\subsection{Human and LLM Evaluation}

\noindent \textbf{Human Participants}
We recruit 128 native English speakers via Prolific. All participants report being born and raised speaking English and having no history of neurological or cognitive conditions affecting language comprehension. Two attention-check questions are included, and participants who fail either check are excluded. 120 participants were retained after applying the exclusion criteria (see Appendix~\ref{app:human} for details).

Both human participants and LLMs receive identical instructions. The full instructions and prompts are reported in Appendix~\ref{app:stimuli} and Appendix~\ref{app:prompt design}.

\paragraph{Experimental Procedure}
Participants complete a presupposition judgment task consisting of the same 90 conditional sentences of the form $\textit{If } A, B_p$. They are instructed to assume that the speaker is honest, reliable, and cooperative, and that relevant background assumptions are shared. For each item, participants rate how likely the presupposition $p$ is to be true on a 0--7 Likert scale (0 = very unlikely, 7 = very likely). Examples of this task are provided in Appendix~\ref{app:B2}, Table~\ref{tab:main_study_examples}.

The experiment follows a between-participants design with two conditions: \emph{without-context} and \emph{with-context}. All participants in both conditions answered the same 90 items, however, in the with-context condition, each item additionally included a brief identifying background description (e.g., the person being discussed is from Toronto). Providing minimal identifying context allows us to show how background information modulates perceived evidential reliability. This is consistent with prior work showing that contextual cues influence how speakers evaluate the strength of testimonial and inferential evidence~\cite{lesage2015reliability}. The rationale for this manipulation is based on psycholinguistic
work on context-driven interpretation~\cite{Crain_Steedman_1985}. In the absence of any contextual framing, participants must
implicitly decide which situation or possible world they are meant to reason about, leaving many parameters of the
interpretive problem underspecified. Even a minimal context
line, such as identifying where an individual is from, serves
two related functions. First, it fixes parameters that would
otherwise remain open (e.g., the individual's background
properties), and it increases the salience of those
parameters, potentially triggering relevance-based reasoning about whether the contextual information bears on the target presupposition~\cite{Sperber1995-SPER}. These are genuine
contextual changes, not trivial additions, and our design allows us to examine how much such changes actually shift presupposition judgments in both humans and LLMs.

To reduce fatigue in the longer with-context condition, two mandatory 8-second pauses were inserted near the beginning and end of the task. All items were presented in randomized order, and participants completed the task in a single session.

\paragraph{LLM Setup and Prompting}
We evaluate four LLMs: GPT-5, Gemini-2.5-flash, Llama-3.1-8B-Instruct, and Qwen2.5-7B-Instruct. Model versions, access methods, and generation parameters are reported in Appendix \ref{model cost}, and the full decoding configuration in Appendix~\ref{sec:inference-setup}. Each model receives the same instructions and stimuli as human participants. All models are instruction-tuned versions, denoted as ``IT''. Details are in Appendix \ref{app:prompt design}.

Each model is presented with the same 90 items under both experimental conditions. Models produce (i) a likelihood judgment on the same 0–7 Likert scale and (ii) an output of CoT reasoning steps. This yields parallel human and model judgments and corresponding reasoning traces.

\subsection{LLM-as-a-Judge Framework}
Because presupposition inference relies on pragmatic reasoning, evaluating final judgments, made by LLMs, alone is insufficient. We therefore adopt an LLM-as-a-Judge framework inspired by \citet{lee-etal-2025-checkeval} and \citet{yu2025think}, which enables structured assessment of reasoning quality of the models using theory-driven criteria.

\noindent \textbf{Checklist Design}
We design a checklist that decomposes pragmatic competence into explicit yes/no questions. Seed questions are created by a linguistics expert based on theories of presupposition accommodation, conditional semantics, and pragmatic reasoning, as well as coherence and consistency criteria~\cite{wang-etal-2020-asking, fabbri-etal-2021-summeval}. These questions are grouped into four dimensions: logical accuracy, presupposition handling, pragmatic criteria, and coherence. For the with-context condition, we add a fifth dimension that addresses context integration.

Each dimension is further divided into sub-dimensions targeting specific reasoning behaviors, based on established work in discourse and dynamic semantics, presupposition projection, and pragmatic reasoning~\cite{grice1975logic,Heim-1983, KampReyle1993, beaver2001presupposition}. Accuracy evaluates whether the conditional structure is correctly interpreted, presupposition handling assesses identification of presupposition triggers and the distinction between presupposition and entailment, pragmatic criteria examine cooperativity, relevance, and common ground, and coherence evaluates consistency between numerical judgments and stated reasoning. Table~\ref{tab:theoretical_framework} summarizes these dimensions, their theoretical bases, and representative evaluation criteria.

To expand coverage, we use \texttt{Llama-3.1-8B-Instruct} to diversify and augment the seed questions and then further refine them through manual elaboration. The same model is used to filter redundant items. All questions are binary, with “yes” indicating alignment with theoretical expectations. The final checklist is reviewed by a linguist to ensure theoretical validity and non-redundancy.

The resulting checklist contains 59 questions for the with-context condition (5 dimensions) and 52 questions for the without-context condition (4 dimensions). A sample is provided in the Appendix~\ref{app:checklist}.\footnote{The full checklist is available in our GitHub repository.} Thus, we define the total criteria space as:
\[
N = \underbrace{|M|}_{\text{Models}} \times \Bigg( \underbrace{|I_{\text{wc}}| \times |\mathcal{K}_{\text{wc}}|}_{\text{With-Context}} + \underbrace{|I_{\text{nc}}| \times |\mathcal{K}_{\text{nc}}|}_{\text{Without-Context}} \Bigg)
\]
where \(|M|\) is the number of models, \(|I_{\text{wc}}|\) and \(|I_{\text{nc}}|\) are the numbers of items in the with-context and without-context conditions, respectively, and \(|\mathcal{K}_{\text{wc}}|\) and \(|\mathcal{K}_{\text{nc}}|\) are the corresponding numbers of checklist questions.

\paragraph{Judge Model Configuration} 
Given the scale of the model checkpoints and the complexity of pragmatic reasoning tasks, we employ an LLM-as-a-Judge framework to enable scalable and consistent evaluation of model reasoning. The judge evaluates whether each reasoning trace satisfies the pragmatic constraints specified by the checklist.

We use \texttt{claude-haiku-4} as the judge model, based on preliminary experiments indicating strong performance on the recent fact-checking benchmark \citep{theologitis2026claimdb}, while also mitigating potential bias arising from shared training corpora within the same provider. For each evaluation instance, the judge receives an input tuple:
\[
\mathcal{I} = (c, s, \tau, r_{\text{CoT}})
\]
where $c$ is the discourse context (or $\emptyset$ in the without-context condition), $s$ is the conditional sentence, $\tau$ is the target presupposition, and $r_{\text{CoT}}$ is the reasoning trace generated by the evaluated model.

\paragraph{Evaluation Procedure} 
A prompt template $P$ presents $(c, s)$ and instructs models to act as pragmatic listeners (see Appendix~\ref{app:prompt design} for the prompts). Each model first produces an explicit reasoning trace and then provides a final likelihood judgment. 
Finally, we deploy the judge model configured in the previous section. For a given model response $r$, the judge iterates through the condition-specific checklist $\mathcal{K}$. For each criterion $\kappa_j \in \mathcal{K}$, the judge model $J$ produces a binary decision $J\big((c, s, \tau, r), \kappa_j\big)$, where 1 indicates that the response satisfies the criterion and 0 otherwise. The final score ($S$) for that response is computed as the average compliance across all criteria:
\[
S = \frac{1}{|\mathcal{K}|} \sum_{j=1}^{|\mathcal{K}|} J\big((c, s, \tau, r), \kappa_j\big), J(\cdot) \in \{0, 1\}
\]
This procedure yields both aggregate performance metrics and granular, dimension-level diagnostics. To assess the reliability of this automated evaluation, we conduct a human validation study on a subset of the judged outputs below.

\paragraph{Human Validation} Two PhD students in linguistics independently evaluated a randomly sampled 5\% of the outputs (36 items, 1,992 binary judgments), drawn with equal representation across models and conditions. Inter-annotator agreement was 89\%, (exact match accuracy) and agreement with the judge model was 79.46\%.

\section{Experiments and Results}

\subsection{Norming Results}

\begin{figure}[t]
    \centering

    \includegraphics[width=0.80\linewidth]{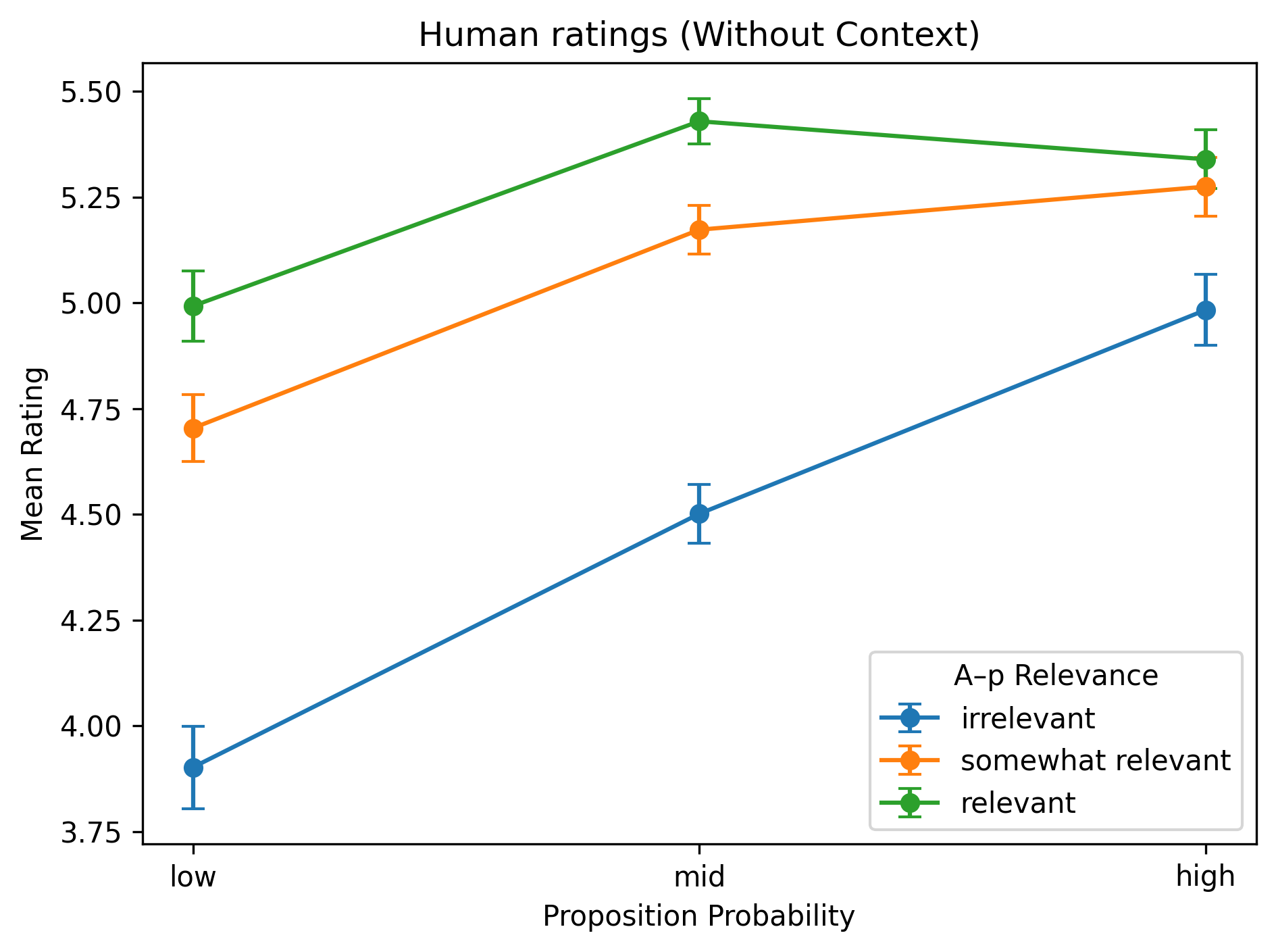}

    \vspace{0.15cm}

    \includegraphics[width=0.80\linewidth]{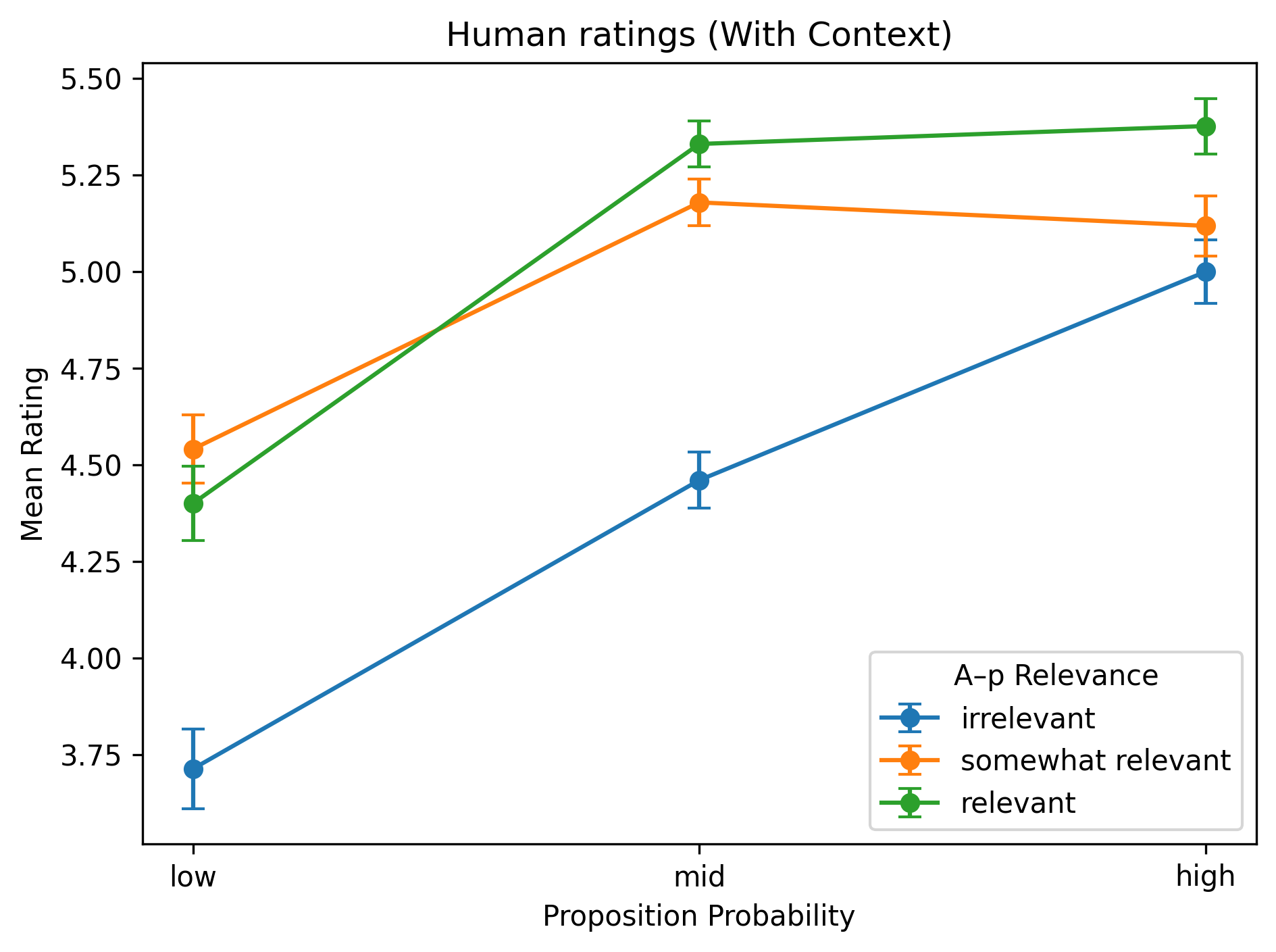}

    \caption{Human mean Likert scores in the main presupposition judgment experiment, across participants and items as a function of proposition probability (low, mid, high) and antecedent–presupposition ($A-p$) relevance (relevant, somewhat relevant, irrelevant). Top: without-context condition; bottom: with-context condition. Error bars indicate standard errors.}
    \label{fig:human_interaction}
    \vspace{-3ex}
\end{figure}

The norming study results closely matched our intended low, mid, and high probability conditions. Mean ratings increased monotonically across relevance levels, with low probability items receiving lower ratings (M = 2.76), mid probability items intermediate ratings (M = 4.38), and high probability items the highest ratings (M = 5.52).

We analyzed the results using a linear mixed-effects model~\cite{baayen2008mixed} with expected probability level (low, mid, high) as a fixed effect and random intercepts for participants and items. This approach allows us to account for repeated measurements as well as variability across participants and items, making it appropriate for our experimental design. The model showed a significant effect of probability level, with ratings increasing from low to mid probability ($\beta = 1.62$, $p < .001$) and from low to high probability ($\beta = 2.75$, $p < .001$). These results confirm that the norming study can successfully be used as the basis for graded \emph{$A$-$p$} probability, as intended.

\subsection{Human Performance}

\setlength{\tabcolsep}{3pt}
\begin{table}[t!]
\centering
\small 
\renewcommand{\arraystretch}{1.1} 
\begin{tabular}{llccc}
\toprule
& & \multicolumn{3}{c}{\textbf{Probability Level (0--7)}} \\
\cmidrule(lr){3-5}
\textbf{Model} & \textbf{Type} & \textbf{Rel.} & \textbf{s/w Rel.} & \textbf{Irrel.} \\
\midrule
\multicolumn{5}{l}{\textit{\textbf{Closed Source}}} \\
\midrule
\multirow{2}{*}{Gemini-2.5-flash} 
 & \cellcolor{ctxgray}w Context & \cellcolor{ctxgray}6.43 & \cellcolor{ctxgray}6.57 & \cellcolor{ctxgray}6.87 \\
 & w/o Context & 6.60 & 6.53 & 6.93 \\
\addlinespace
\multirow{2}{*}{GPT-5} 
 & \cellcolor{ctxgray}w Context & \cellcolor{ctxgray}5.43 & \cellcolor{ctxgray}5.87 & \cellcolor{ctxgray}6.17 \\
 & w/o Context & 5.83 & 5.93 & 6.53 \\
\midrule
\multicolumn{5}{l}{\textit{\textbf{Open Source}}} \\
\midrule
\multirow{2}{*}{Llama3.1-8B-IT} 
 & \cellcolor{ctxgray}w Context & \cellcolor{ctxgray}5.07 & \cellcolor{ctxgray}4.70 & \cellcolor{ctxgray}4.30 \\
 & w/o Context & 5.60 & 4.83 & 5.17 \\
\addlinespace
\multirow{2}{*}{Qwen2.5-7B-IT} 
 & \cellcolor{ctxgray}w Context & \cellcolor{ctxgray}6.00 & \cellcolor{ctxgray}5.80 & \cellcolor{ctxgray}5.73 \\
 & w/o Context & 6.37 & 6.20 & 6.03 \\
\bottomrule
\end{tabular}
\caption{Mean likelihood ratings (0--7 scale) produced by each model across $A$-$p$ relevance (relevant, somewhat relevant, irrelevant) under with-context and without-context prompting conditions. Shaded rows indicate the with-context condition. IT abbreviates Instruct.}

\label{tab:model_performance}
\vspace{-3ex}
\end{table}

\noindent \textbf{A-p Relevance and Probability}
We analyzed human judgments using linear mixed-effects models with fixed effects of proposition probability (low, mid, high), \emph{$A$-$p$} relevance (relevant, somewhat relevant, irrelevant), and their interaction, with random intercepts for participants. Proposition probability was treated as a three-level categorical factor rather than a continuous predictor to preserve alignment with the experimental design, in which items were constructed to instantiate discrete probability levels, and to facilitate interpretable comparison across conditions. Full model output, including the model formula and fit statistics, is reported in Appendix~\ref{app:LMM_Results_human} (Table~\ref{tab:lmm_human_combined}). Separate models were fitted for the without-context and with-context conditions. Figure~\ref{fig:human_interaction} summarizes the resulting response patterns.
In the without-context condition, items with high proposition probability and high $A$-$p$ relevance (e.g., having a smartphone in \textit{If Alex is a college student, he will use his smartphone to take notes}) received the highest ratings (M $\approx$ 5.34). Low-probability items (e.g., \textit{having a Rolex}) were rated significantly lower, while mid-probability items (e.g., \textit{having a brother}) were not reliably distinguished from high-probability ones. Similarly, items with irrelevant antecedents (e.g., \textit{liking coffee} ($A$) and \textit{having a wetsuit} ($p$)) received substantially lower ratings, whereas somewhat relevant items (e.g., \textit{liking swimming} and \textit{having a wetsuit}) were treated similarly to highly relevant ones. A significant interaction showed that low and mid probability led to especially strong penalties when $A$ and $p$ were irrelevant 
(low $\times$ irrelevant: $\beta = -0.73$, $z = -5.47$, $p < .001$; 
mid $\times$ irrelevant: $\beta = -0.57$, $z = -4.61$, $p < .001$; 
see Table~\ref{tab:lmm_human_combined}). This indicates that participants combined probability and relevance in a non-additive manner, consistent with probabilistic accounts that predict interactive effects between antecedent relevance and prior likelihood. Notably, when $A$ and $p$ were irrelevant, ratings tracked proposition probability closely, suggesting that participants fell back on prior likelihood in the absence of a meaningful antecedent-presupposition relation. In contrast, even partial $A-p$ relevance produced a marked boost in ratings, indicating that the relevance cue acts less as a graded scalar and more as a threshold.

In the with-context condition, baseline ratings for highly probable and highly relevant items remained similar (M $\approx$ 5.38), indicating no overall shift in response levels. However, participants used both cues more consistently. Low probability propositions showed a stronger decrease in ratings, and both somewhat relevant and irrelevant items were rated significantly lower than highly relevant ones. This suggests that contextual information encouraged more graded integration of probability and relevance.

\noindent \textbf{Context vs.\ No-Context Effects} Across both conditions, baseline judgments for highly probable and highly relevant items are nearly identical, suggesting that minimal context does not affect default interpretations in clear cases. However, context modulates how participants use probability and relevance cues.
Without context, participants primarily distinguish between relevant and irrelevant items, with irrelevant $A$--$p$ relations acting as a strong gating factor and probability playing a relatively modest role. In contrast, with context, participants show a more differentiated use of both cues, consistent with the interaction effects observed in the mixed-effects analysis.
For illustration, consider the item \textit{If Jack is a scuba diver, he will bring his wetsuit}, which exemplifies the overall pattern visible in Figure~\ref{fig:human_interaction}. Statistical conclusions are based on the full model reported above.


\begin{figure}[t]
    \centering

    \includegraphics[width=1.1\linewidth]{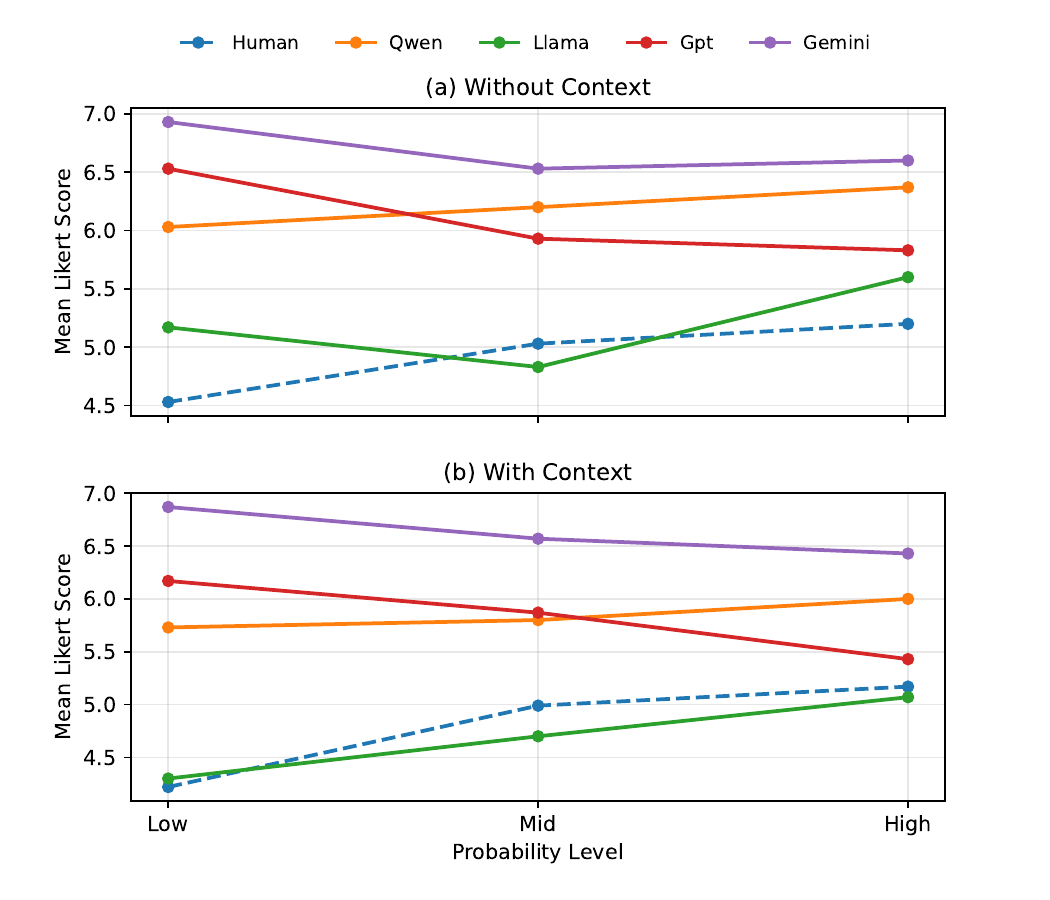}

    \caption{Mean Likert scores for human participants and each LLM in the main presupposition judgment experiment, as a function of proposition probability (low, mid, high) and $A-p$ relevance, under without-context (top) and with-context (bottom) conditions.}

    \label{fig:human_llm_mean}
    \vspace{-3ex}
\end{figure}

\subsection{Human-LLM Comparison}
\begin{table*}[t]
\centering
\small
\renewcommand{\arraystretch}{1.25} 
\setlength{\tabcolsep}{4pt}       

\begin{tabular}{l cc cc | cc cc}
\toprule

& \multicolumn{4}{c}{\textbf{Proprietary Models}} 
& \multicolumn{4}{c}{\textbf{Open-source Models}} \\
\cmidrule(lr){2-5} \cmidrule(lr){6-9}

& \multicolumn{2}{c}{\textbf{Gemini-2.5-flash}} & \multicolumn{2}{c}{\textbf{GPT-5}} 
& \multicolumn{2}{c}{\textbf{Llama3.1-8B-IT}} & \multicolumn{2}{c}{\textbf{Qwen2.5-7B-IT}} \\
\cmidrule(lr){2-3} \cmidrule(lr){4-5} \cmidrule(lr){6-7} \cmidrule(lr){8-9}

\textbf{Category} & 
\cellcolor{ctxgray}\textbf{With Ctx (\%)} & \textbf{$\Delta$} & 
\cellcolor{ctxgray}\textbf{With Ctx (\%)} & \textbf{$\Delta$} & 
\cellcolor{ctxgray}\textbf{With Ctx (\%)} & \textbf{$\Delta$} & 
\cellcolor{ctxgray}\textbf{With Ctx (\%)} & \textbf{$\Delta$} \\
\midrule

Accuracy & 
\cellcolor{ctxgray}61.39 & \diffdown{6.39} & 
\cellcolor{ctxgray}60.74 & \diffdown{2.50} & 
\cellcolor{ctxgray}16.67 & \diffdown{13.98} & 
\cellcolor{ctxgray}22.96 & \diffdown{13.80} \\

Coherence & 
\cellcolor{ctxgray}69.41 & \diffup{5.30} & 
\cellcolor{ctxgray}75.11 & \diffup{4.56} & 
\cellcolor{ctxgray}62.81 & \diffup{6.15} & 
\cellcolor{ctxgray}58.74 & \diffup{5.63} \\

Pragmatic & 
\cellcolor{ctxgray}82.84 & \diffdown{5.63} & 
\cellcolor{ctxgray}78.89 & \diffdown{3.19} & 
\cellcolor{ctxgray}54.94 & \diffup{5.63} & 
\cellcolor{ctxgray}56.42 & \diffdown{3.86} \\

Presupposition & 
\cellcolor{ctxgray}61.11 & \diffdown{12.93} & 
\cellcolor{ctxgray}71.02 & \diffup{0.51} & 
\cellcolor{ctxgray}49.35 & \diffdown{8.32} & 
\cellcolor{ctxgray}42.04 & \diffdown{15.44} \\

\rowcolor{ctxblue} 
\textit{Context Util.} & 
30.10 & \nodiff & 
28.18 & \nodiff & 
14.75 & \nodiff & 
12.42 & \nodiff \\

\textbf{TOTAL} & 
\cellcolor{ctxgray}\textbf{60.81} & \textbf{\diffdown{11.79}} & 
\cellcolor{ctxgray}\textbf{63.18} & \textbf{\diffdown{7.47}} & 
\cellcolor{ctxgray}\textbf{40.53} & \textbf{\diffdown{7.36}} & 
\cellcolor{ctxgray}\textbf{39.08} & \textbf{\diffdown{11.82}} \\
\bottomrule

\end{tabular}
\caption{The $\Delta$ columns indicate the performance gap relative to the \textit{Without Context} condition (\textcolor{diffpos}{Green $\uparrow$} indicates \textit{With Context} scored higher; 
\textcolor{diffneg}{Red $\downarrow$} indicates \textit{With Context} scored lower). IT abbreviates Instruct.}
\label{tab:full_comparison}
\vspace{-3ex}
\end{table*}

We assessed the alignment between human and model judgments using Spearman rank correlations and mean absolute error (MAE). Table~\ref{tab:model_performance} summarizes the average model ratings across $A$-$p$ relevance and context conditions. Figure~\ref{fig:human_llm_mean} provides a visual comparison of human and model responses across probability levels under both contextual conditions. At the item level, models differed substantially in their degree of alignment with human judgments. Qwen-2.5-7B-IT showed the strongest and most reliable correlations in both conditions (without context: $\rho = 0.25$, $p = .016$; with context: $\rho = 0.38$, $p < .001$), very close to the human's ranking of items. Llama3.1-8B-Instruct also showed consistent moderate alignment (without context: $\rho = 0.21$, $p = .047$; with context: $\rho = 0.30$, $p = .004$). In contrast, Gemini-2.5-flash showed meaningful alignment only when context was provided ($\rho = 0.26$, $p = .013$). GPT-5 did not show significant overall correlations in either condition.

\begin{figure}[h]
    \centering
    \includegraphics[width=\columnwidth]{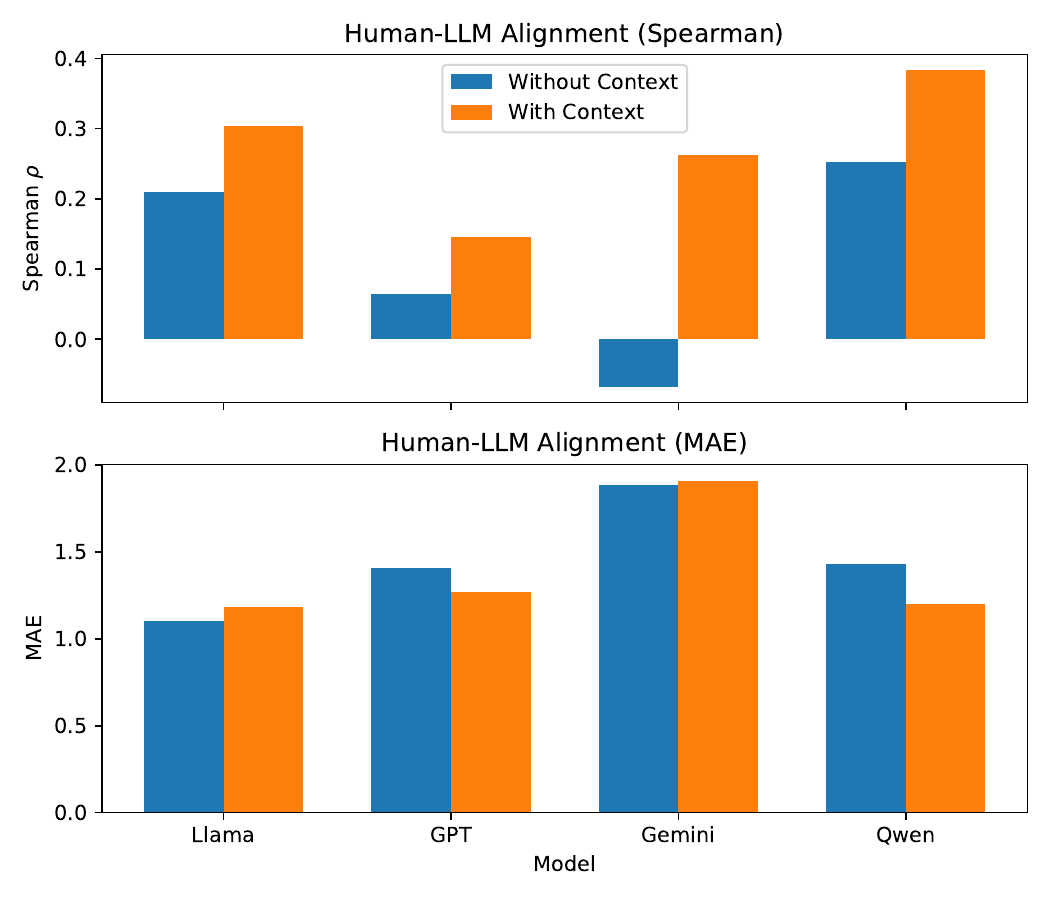}
    \caption{Human-LLM alignment measured using Spearman’s rank correlation ($\rho$) and mean absolute error (MAE) under with-context and without-context conditions. Higher Spearman values and lower MAE indicate closer alignment between model predictions and human mean Likert scores.}
    \label{fig:human_llm_alignment}
    \vspace{-3ex}
\end{figure}

These patterns were mirrored in the MAE results (Figure~\ref{fig:human_llm_alignment}). Llama3.1-8B-Instruct achieved the lowest overall error (MAE = 1.14), followed by Qwen2.5-7B-IT (1.32) and GPT-5 (1.34), whereas Gemini-2.5-flash showed the largest deviation from human judgments (1.90). For GPT-5 and Qwen2.5-7B-IT, MAE was slightly lower in the with-context condition, and Llama3.1-8B-Instruct and Gemini-2.5-flash showed comparable error levels across conditions. Overall, Qwen2.5-7B-IT and Llama3.1-8B-Instruct demonstrated the strongest and most reliable correspondence with human presupposition judgments based on Likert scale. Gemini-2.5-flash’s performance depended on contextual information, and GPT-5 showed limited alignment across conditions.

To examine whether these quantitative patterns are reflected in the models’ reasoning, we next used the LLM-as-a-Judge framework to evaluate whether their reasoning steps are consistent with the scores they assigned on the Likert scale.

\subsection{Reasoning Analysis}

To examine whether model explanations align with their assigned Likert scores, we analyze checklist-based evaluation results produced by the LLM-as-a-Judge framework. The breakdown of reasoning scores is presented in Table \ref{tab:full_comparison}.

\paragraph{Stronger Theoretical Alignment in Larger Instruction-Tuned Models}
Across evaluation dimensions, larger and more heavily instruction-tuned models in our comparison (Gemini-2.5-flash and GPT-5) achieve higher checklist compliance than smaller open-source models. In the with-context condition, proprietary models achieve total scores above 60\%, while open-source models remain near 40\%. This gap is particularly pronounced in the Accuracy and Presupposition dimensions.  

These results align with scaling-law observations \citep{ruan2024observational} and the effects of instruction tuning \citep{jiang2024instruction}. Larger instruction-tuned models appear better at maintaining structured reasoning and tracking presuppositional dependencies across multi-step explanations. Nevertheless, even these models exhibit limitations in graded presupposition reasoning. For instance, GPT-5 assigns the same Likert rating (6) to both \textit{If Jack is a scuba diver, he'll bring his wetsuit} and \textit{If Sara likes coffee, she’ll bring her wetsuit}. In its explanation, the model emphasizes the possessive trigger (“his wetsuit”) and general ownership knowledge without evaluating whether the antecedent increases $Pr(p \mid A, c)$ relative to $Pr(p \mid c)$. This pattern suggests reliance on lexical cues rather than genuine sensitivity to presupposition relevance.

\paragraph{Context Improves Checklist Performance} Providing minimal identifying context generally increases checklist compliance. In particular, Llama-3.1-8B-IT and Qwen2.5-7B-IT show notable gains in the Accuracy and Presupposition dimensions under the with-context condition. Proprietary models also benefit in some dimensions, although improvements are more modest.  

This trend aligns with prior work \citep{lee2025reasoning}, which finds that abstract context and world knowledge do not shift baseline interpretations but increase sensitivity to probabilistic structure. For models, however, context often introduces reasoning noise, as attempts to incorporate contextual details frequently fail to align consistently with conditional semantics, leading to reduced coherence compliance.

\paragraph{Dissociation Between Theoretical Reasoning and Probabilistic Presupposition}  
We observe a clear dissociation between behavioral alignment (Likert ratings) and explicit reasoning quality (checklist compliance). As shown in Table \ref{tab:model_performance} and Figure \ref{fig:human_llm_alignment}, Qwen2.5-7B-IT is most closely aligned with human Likert ratings across all conditions. However, the reasoning results in Table \ref{tab:full_comparison} reveal a paradox: Qwen2.5-7B-IT shows the lowest average compliance with our theory-informed checklist.

This pattern suggests that smaller models approximate human-like response distributions by relying on surface-level lexical associations and distributional world knowledge, without consistently implementing the formal pragmatic constraints underlying the proviso problem. In contrast, proprietary models generate more structured and theory-consistent reasoning steps, yet their final Likert ratings deviate more from human judgments. We interpret this dissociation as suggesting that behavioral similarity does not necessarily reflect stable pragmatic competence. High alignment in ratings may arise from probabilistic pattern matching over training data, rather than explicit modeling of relevance relations between antecedent and presupposition. At the same time, we acknowledge that checklist compliance may partly reflect verbalization quality rather than reasoning depth, given evidence that chain-of-thought traces can function as post-hoc rationalizations in instruction-tuned models~\citep{turpin_etal}. The dissociation is therefore consistent with either a genuine competence gap in smaller models or a verbalization advantage in larger ones, and disentangling these interpretations is an important direction for future work.

\section{Conclusion}

This study presents an empirical comparison of human and LLM presupposition judgments in conditional sentences grounded in pragmatic theory. Using a norming study and parallel behavioral experiments, we show that human presupposition judgments are guided by intra-sentential relationships (e.g., between antecedent and presupposition), proposition probability, and contextual information. In contrast, LLMs show varying degrees of alignment, with Qwen2.5-7B-IT closest to human patterns, but generally struggle with graded presupposition accommodation.
Our LLM-as-a-Judge analysis reveals a dissociation between behavioral alignment and reasoning quality. Models that best match human ratings often fail to satisfy theory-derived pragmatic criteria, while models that better meet theoretical standards produce less human-like judgments. This suggests that current performance may reflect surface-level pattern matching rather than stable pragmatic understanding.

Future work should extend this framework to additional presupposition triggers and more complex constructions. Theory-driven evaluation remains crucial for characterizing semantic and pragmatic reasoning in language models and improving human–model alignment.

\section*{Limitations}

Our study focuses on conditional sentences of a specific structural form ($\textit{If } A, B_p$) and investigates presupposition projection through possessive pronoun triggers (e.g., \textit{his}, \textit{her}, \textit{their}). We focus on possessive pronouns because they introduce simple existential presuppositions that are well studied in the projection literature (e.g. \citealt{Karttunen-1973}, \citealt{Heim-1983}) and allow controlled testing of the $A$-$p$ relevance without additional lexical confounding. As a result, our findings may not generalize to other types of conditionals, presupposition triggers (e.g., factives, change-of-state verbs), or more complex embedding environments.

Moreover, our human data is collected from English speaking participants in a limited set of countries, which may restrict cross-linguistic and cross-cultural generalizability. Although we validated a subset of the LLM-as-a-Judge outputs with human annotators due to practical cost considerations, future work could expand this validation to additional samples to further support and refine the evaluation framework.

Finally, the LLM-as-a-Judge checklist operationalizes pragmatic competence through satisfaction-theoretic principles, which may not fully capture the heuristic and probabilistic nature of human presupposition accommodation. The observed dissociation between behavioral alignment and checklist compliance is therefore potentially ambiguous: it may reflect a genuine competence gap in models, or it may reflect a mismatch between the formal criteria encoded in the checklist and the cognitive processes that actually drive human judgments. Future work could complement theory-driven evaluation with process-level measures to disentangle these interpretations.

\bibliography{custom}


\appendix

\section{Human Study Materials}
\label{app:human}

\subsection{Norming}

The norming study included 30 native English speakers (18 men, 12 women), aged between 25 and over 65. Participants were primarily based in the United States ($n = 22$), with other participants from the United Kingdom ($n = 4$) and Canada ($n = 4$). Educational backgrounds ranged from high school to graduate degrees.

Participants were recruited through an online platform (Prolific) and were required to be native or near-native English speakers who were raised speaking English from birth. Eligibility criteria included current residence in the United States, the United Kingdom, or Canada, age between 18 and 99, and a high prior approval rate on Prolific (95–100\%). Participants were screened for language background, nationality, and country of birth to ensure alignment with the target population. Individuals reporting a history of neurological, cognitive, or language-related conditions were excluded. All participants completed the study in a single session and received monetary compensation at an hourly rate of £12. All participants provided informed consent electronically, on the platform, prior to participation and were informed that their anonymized responses would be used for research purposes. The same recruitment requirements, screening criteria, and consent procedures are applied in the main experiment. The study protocol for both norming and presupposition judgment task is approved by the institutional Research Ethics Board.

In this task, participants rated 120 short sentences on a 0–7 Likert scale (0 = very unlikely, 7 = very likely) indicating how likely each statement was to be true in the real world. Two attention checks were included. The sentences varied in expected likelihood. For example, a baseline item asked, \textit{How likely is it that someone chosen at random from your hometown has a stamp collection?}, High-probability items added a strongly relevant modifier (e.g., ``someone who is interested in postal history''), mid-probability items included a moderately relevant modifier (e.g., ``a retired postman''), and low-probability items included an unrelated modifier (e.g., ``someone who is vegetarian''). The expected completion time was approximately 35–40 minutes the task was designed to be finished in one sitting (median completion time: 26.8 minutes).

\subsection{Main Presupposition Judgment Study}

The main study consisted of two independent groups corresponding to the with-context and without-context conditions. Different participants were recruited for each condition to avoid carryover effects.
In total, data from 128 participants were collected. After excluding eight participants who failed attention checks, data from 120 participants (60 per condition) were retained for analysis. The final sample consisted of 55 women, 64 men, and one non-binary participant, residing in the United States ($n = 73$), the United Kingdom ($n = 29$), and Canada ($n = 18$), aged between 25 and over 65. Educational backgrounds ranged from high school to graduate degrees.

Participants completed 90 items, each consisting of a conditional statement and a corresponding target statement, and rated the likelihood of the target statement on a 0–7 Likert scale. In the with-context condition, participants were additionally provided with brief identifying background information about the individual described in each item.

The median completion time was 23.5 minutes for the with-context condition and 19.9 minutes for the without-context condition.

Participants were compensated at an hourly rate of approximately £10. To reduce fatigue and cognitive load, two mandatory 8-second pauses were included in the with-context study.

Since human participants and LLMs received largely identical instructions, except that LLMs were additionally asked to provide step-by-step reasoning, we do not reproduce them here. The full instructions and prompts are reported in the following section of the Appendix.

\section{Example Stimuli}
\label{app:stimuli}

This section presents representative examples of stimuli used in the norming study and the main presupposition judgment experiment. All stimuli were presented in randomized order during the experiments, and probability and \emph{$A$-$p$} relevance labels are shown here for explanatory purposes only.

\subsection{Norming Study Items}

The norming study was conducted to elicit high-, mid-, and low-probability propositions and to confirm, through human judgments, that these propositions exhibit the intended probability levels in real-world contexts. Table~\ref{tab:norming_examples} presents representative examples from the norming study. Each item consists of a baseline condition and three conditional variants corresponding to high, mid, and low probability levels. The resulting ratings were used to construct the stimulus set for the main presupposition judgment experiment.

\begin{table*}[t]
\centering
\small
\renewcommand{\arraystretch}{1.2}

\begin{tabular}{p{2.5cm} p{12.5cm}}
\toprule
\textbf{Condition} & \textbf{Item} \\
\midrule

\multicolumn{2}{l}{\textbf{Low-Probability Example: Stamp Collection}} \\
\midrule

Scenario & Someone from your hometown / having a stamp collection \\

Baseline &
How likely is it that someone chosen at random from your hometown has a stamp collection? \\

High &
How likely is it that someone chosen at random from your hometown who is interested in postal history has a stamp collection? \\

Mid &
How likely is it that someone chosen at random from your hometown who is a retired postman has a stamp collection? \\

Low &
How likely is it that someone chosen at random from your hometown who is vegetarian has a stamp collection? \\

\midrule
\multicolumn{2}{l}{\textbf{Mid-Probability Example: Having Grandchildren}} \\
\midrule

Scenario & Someone from your neighborhood / having grandchildren \\

Baseline &
How likely is it that someone chosen at random from your neighborhood has grandchildren? \\

High &
How likely is it that someone chosen at random from your neighborhood who has adult children has grandchildren? \\

Mid &
How likely is it that someone chosen at random from your neighborhood who is over 60 has grandchildren? \\

Low &
How likely is it that someone chosen at random from your neighborhood who is left-handed has grandchildren? \\

\midrule
\multicolumn{2}{l}{\textbf{High-Probability Example: Credit Cards}} \\
\midrule

Scenario & Someone at a coffee shop / having credit cards \\

Baseline &
How likely is it that someone chosen at random at a coffee shop has credit cards? \\

High &
How likely is it that someone chosen at random at a coffee shop who travels frequently has credit cards? \\

Mid &
How likely is it that someone chosen at random at a coffee shop who works full-time has credit cards? \\

Low &
How likely is it that someone chosen at random at a coffee shop who drinks tea has credit cards? \\

\bottomrule
\end{tabular}

\caption{Representative examples from the norming study showing baseline and high-, mid-, and low-relevance conditions for the propositions.}
\label{tab:norming_examples}
\end{table*}

\subsection{Main Presupposition Judgment Items}
\label{app:B2}
The main presupposition judgment task is designed to investigate how participants integrate antecedent-presupposition relevance and contextual information when interpreting conditional sentences. Each item consists of a background description, a conditional statement, and a target presupposition, and is presented under either with-context or without-context conditions. Table~\ref{tab:main_study_examples} presents representative examples of the experimental stimuli.

\newcolumntype{L}[1]{>{\raggedright\arraybackslash}p{#1}}
\begin{table*}[t]
\centering
\footnotesize
\renewcommand{\arraystretch}{1.25}
\setlength{\tabcolsep}{3pt}

\begin{tabular}{L{5cm} L{7.4cm} L{3.4cm}}
\toprule
\textbf{Background} & \textbf{Statement 1 (Conditional)} & \textbf{Target} \\
\midrule

\multicolumn{3}{l}{\textbf{having an Instagram account (High-Probability Example)}} \\
\midrule

\rowcolor{ctxgray}
Alex works at the airport in your town.
&
\textbf{Rel:}  If Alex uses social media, he will post a photo to his Instagram account. &
Alex has an Instagram account. \\

Maya works at the airport in your town. &
\textbf{S-Rel:} If Maya is over 50, she will check her Instagram account. &
Maya has an Instagram account.
\\

\rowcolor{ctxgray}
Daniel works at the airport in your town. &
\textbf{Irrel:} If Daniel wears blue, he will log out of his Instagram account.
&
Daniel has an Instagram account.\\

\midrule

\multicolumn{3}{l}{\textbf{having a brother (Mid-Probability Example)}} \\
\midrule

\rowcolor{ctxgray}
John works out at the downtown gym. &
\textbf{Rel:}  If John has siblings, he will call his brother on the weekend. &
John has a brother. \\

Nadine works out at the downtown gym. &
\textbf{S-Rel:} If Nadine has a large family, she will invite her brother. &
Nadine has a brother. \\

\rowcolor{ctxgray}
Nick works out at the downtown gym. &
\textbf{Irrel:}  If Nick is left-handed, he will teach his brother mathematics.&
Nick has a brother. \\

\midrule
\multicolumn{3}{l}{\textbf{Having a Wetsuit (Low-Probability Example)}} \\
\midrule

\rowcolor{ctxgray}
Jack is from Ottawa. &
\textbf{Rel:} If Jack is a scuba diver, he will bring his wetsuit. &
Jack has a wetsuit. \\

Tim is from Ottawa. &
\textbf{S-Rel:} If Tim likes swimming, he will pack his wetsuit. &
Tim has a wetsuit. \\

\rowcolor{ctxgray}
Sara is from Ottawa. &
\textbf{Irrel:} If Sara likes coffee, she will forget her wetsuit. &
Sara has a wetsuit. \\

\bottomrule
\end{tabular}

\caption{Representative examples from the main presupposition judgment task. Each item consists of a background description, a conditional statement, and a target presupposition. Background information was provided only in the with-context condition and omitted in the without-context condition. Probability-level labels were used for analysis and were not shown to participants. The labels Rel, S-Rel, and Irrel indicate levels of antecedent--presupposition relevance (relevant, somewhat relevant, and irrelevant) and were not shown to the participants.}

\label{tab:main_study_examples}
\end{table*}

\section{Linear Mixed-Effects Model Results}
\label{app:LMM_Results_human}

Table~\ref{tab:lmm_human_combined} reports the full fixed-effects output for the linear mixed-effects models fitted to human presupposition judgments in the with-context and without-context conditions, respectively. Both models were specified as follows:

\[
\text{rating} \sim \text{probability} \times \text{relevance} + (1 \mid \text{participant})
\]

In this specification, probability (low, mid, high) and $A-p$ relevance (irrelevant, somewhat relevant, relevant) were treatment-coded factors, with high probability and relevant as the reference levels. Random intercepts were included for participants. Models were fit by restricted maximum likelihood (REML) using the \texttt{MixedLM} implementation in \texttt{statsmodels}. The without-context model included 5,400 observations from 60 participants (log-likelihood $= -10018.32$, residual scale $= 2.29$). The with-context model included 5,400 observations from 60 participants (log-likelihood $= -10240.25$, residual scale $= 2.48$).

\begin{table*}[t]
\centering
\small
\renewcommand{\arraystretch}{1.1}
\setlength{\tabcolsep}{4pt}

\textbf{(A) With-context condition}
\vspace{4pt}

\begin{tabular}{lrrrrc}
\toprule
\textbf{Predictor} & \textbf{$\beta$} & \textbf{SE} & \textbf{$z$} & \textbf{$p$} & \textbf{95\% CI} \\
\midrule
Intercept                                        & 5.377   & 0.159 & 33.82  & $<.001$ & $[5.065,\ 5.689]$   \\
Probability: low                                 & $-$0.977 & 0.099 & $-$9.89 & $<.001$ & $[-1.171,\ -0.783]$ \\
Probability: mid                                 & $-$0.046 & 0.091 & $-$0.51 & .612    & $[-0.225,\ 0.133]$  \\
Relevance: irrelevant                            & $-$0.377 & 0.102 & $-$3.71 & $<.001$ & $[-0.576,\ -0.178]$ \\
Relevance: somewhat relevant                     & $-$0.258 & 0.102 & $-$2.54 & .011    & $[-0.458,\ -0.059]$ \\
Prob.: low $\times$ Rel.: irrelevant             & $-$0.310 & 0.140 & $-$2.22 & .027    & $[-0.584,\ -0.036]$ \\
Prob.: mid $\times$ Rel.: irrelevant             & $-$0.493 & 0.129 & $-$3.82 & $<.001$ & $[-0.747,\ -0.240]$ \\
Prob.: low $\times$ Rel.: somewhat relevant      &  0.399  & 0.140 &  2.86  & .004    & $[0.125,\ 0.673]$   \\
Prob.: mid $\times$ Rel.: somewhat relevant      &  0.107  & 0.129 &  0.83  & .407    & $[-0.146,\ 0.360]$  \\
\midrule
Random intercept variance (Participant)          &  1.206  & 0.145 & \multicolumn{3}{c}{---} \\
\bottomrule
\end{tabular}

\vspace{10pt}

\textbf{(B) Without-context condition}
\vspace{4pt}

\begin{tabular}{lrrrrc}
\toprule
\textbf{Predictor} & \textbf{$\beta$} & \textbf{SE} & \textbf{$z$} & \textbf{$p$} & \textbf{95\% CI} \\
\midrule
Intercept                                        & 5.340   & 0.141 & 37.99  & $<.001$ & $[5.064,\ 5.615]$   \\
Probability: low                                 & $-$0.347 & 0.095 & $-$3.66 & $<.001$ & $[-0.533,\ -0.161]$ \\
Probability: mid                                 &  0.090  & 0.088 &  1.02  & .306    & $[-0.082,\ 0.262]$  \\
Relevance: irrelevant                            & $-$0.356 & 0.098 & $-$3.65 & $<.001$ & $[-0.548,\ -0.165]$ \\
Relevance: somewhat relevant                     & $-$0.065 & 0.098 & $-$0.66 & .509    & $[-0.256,\ 0.127]$  \\
Prob.: low $\times$ Rel.: irrelevant             & $-$0.734 & 0.134 & $-$5.47 & $<.001$ & $[-0.998,\ -0.471]$ \\
Prob.: mid $\times$ Rel.: irrelevant             & $-$0.572 & 0.124 & $-$4.61 & $<.001$ & $[-0.815,\ -0.329]$ \\
Prob.: low $\times$ Rel.: somewhat relevant      & $-$0.224 & 0.134 & $-$1.67 & .095    & $[-0.487,\ 0.039]$  \\
Prob.: mid $\times$ Rel.: somewhat relevant      & $-$0.192 & 0.124 & $-$1.55 & .122    & $[-0.435,\ 0.052]$  \\
\midrule
Random intercept variance (Participant)          &  0.899  & 0.113 & \multicolumn{3}{c}{---} \\
\bottomrule
\end{tabular}

\caption{Linear mixed-effects model results for human presupposition judgments. (A) With-context condition. (B) Without-context condition. Reference levels: Probability = high, Relevance = relevant. Models were fit using REML; each includes 5,400 observations from 60 participants.}
\label{tab:lmm_human_combined}
\end{table*}

\section{Model Sources and Computational Cost}
\label{model cost}
The models evaluated in this paper are obtained from the following sources:
\begin{enumerate}
    \item \textbf{GPT-5} is provided by OpenAI. The corresponding API documentation is available at \url{https://platform.openai.com/docs/models}.
    \item \textbf{Gemini-2.5-flash} is provided by Google Gemini, with API documentation available at \url{https://ai.google.dev/gemini-api/docs}.
    \item \textbf{Claude-haiku-4} is provided by Anthropic. The corresponding API documentation is available at \url{https://platform.claude.com/docs/en/intro}.
    \item \textbf{Qwen2.5-7B-Instruct}\footnote{\url{https://huggingface.co/Qwen/Qwen2.5-7B-Instruct}} and \textbf{Llama3.1-8B-Instruct}\footnote{\url{https://huggingface.co/meta-llama/Llama-3.1-8B-Instruct}} are open-source base model weights obtained from Hugging Face (\url{https://huggingface.co/}).
\end{enumerate}
For large proprietary models (e.g., GPT-5 and Gemini-2.5-Flash), a single run on 360 samples costs approximately \$14 CAD for explanation generation. For the judge model, Claude-Haiku-4, evaluating approximately 40,000 checklist items costs around \$55 CAD. All open-source model evaluations are conducted on a system equipped with two NVIDIA RTX 4090 GPUs (32 GB memory each). Overall, the modest computational requirements demonstrate that the proposed evaluation protocol is accessible to researchers with limited computational resources, while still enabling a comprehensive assessment of state-of-the-art models.

\section{Inference and Decoding Configuration} 
\label{sec:inference-setup} 
All four evaluated LLMs produce a single response per item; no self-consistency or re-sampling is used. 

\paragraph{Open-source Models (Llama-3.1-8B-Instruct, Qwen2.5-7B-Instruct).} 
Run locally with HuggingFace \texttt{transformers} in \texttt{bfloat16} via \texttt{AutoModelForCausalLM.generate} (\texttt{do\_sample=True}, \texttt{temperature}$=0.7$, \texttt{top\_p}$=0.9$, \texttt{max\_new\_tokens}$=1024$; \texttt{top\_k} and \texttt{repetition\_penalty} left at library defaults of $50$ and $1.0$). Prompts use the tokenizer's chat template, are truncated to $4{,}096$ input tokens, and generation stops on the model's default \texttt{eos\_token\_id}; no custom stop strings are set. 

\paragraph{Closed-source Models (GPT-5, Gemini-2.5-Flash).} Accessed through the official OpenAI and Google Generative APIs (\texttt{chat.completions.create} and \texttt{GenerativeModel.generate\_content}, respectively). \texttt{temperature} and \texttt{max\_(output\_)tokens} are set with the same parameter as open-source LMs, matching the condition a typical downstream user encounters. 

\paragraph{Judge Model (Claude-Haiku-4-5).} Called via the Anthropic APIs (\texttt{messages.create}) with \texttt{max\_tokens}$=100$ to accommodate the binary True/False; all other sampling parameters use Anthropic defaults. We release all raw generations and judge decisions alongside the code.

\section{Prompt Design}
\label{app:prompt design}

This section documents the prompt templates used for eliciting likelihood judgments and reasoning from LLMs, as well as the prompts used in the LLM-as-a-Judge evaluation. All task prompts were designed to closely parallel the instructions provided to human participants, with the additional requirement that models produce explicit step-by-step reasoning prior to reporting a final numerical rating.

Figure~\ref{fig:prompt_conditions} presents the system prompts used in the without-context and with-context conditions of the main presupposition judgment task. In both conditions, models were instructed to interpret the speaker as honest and cooperative, to consider relevant background knowledge, and to provide a likelihood rating on a 0--7 Likert scale. The with-context prompt also included identifying background information, parallel to the corresponding human condition.

\begin{figure*}[t]
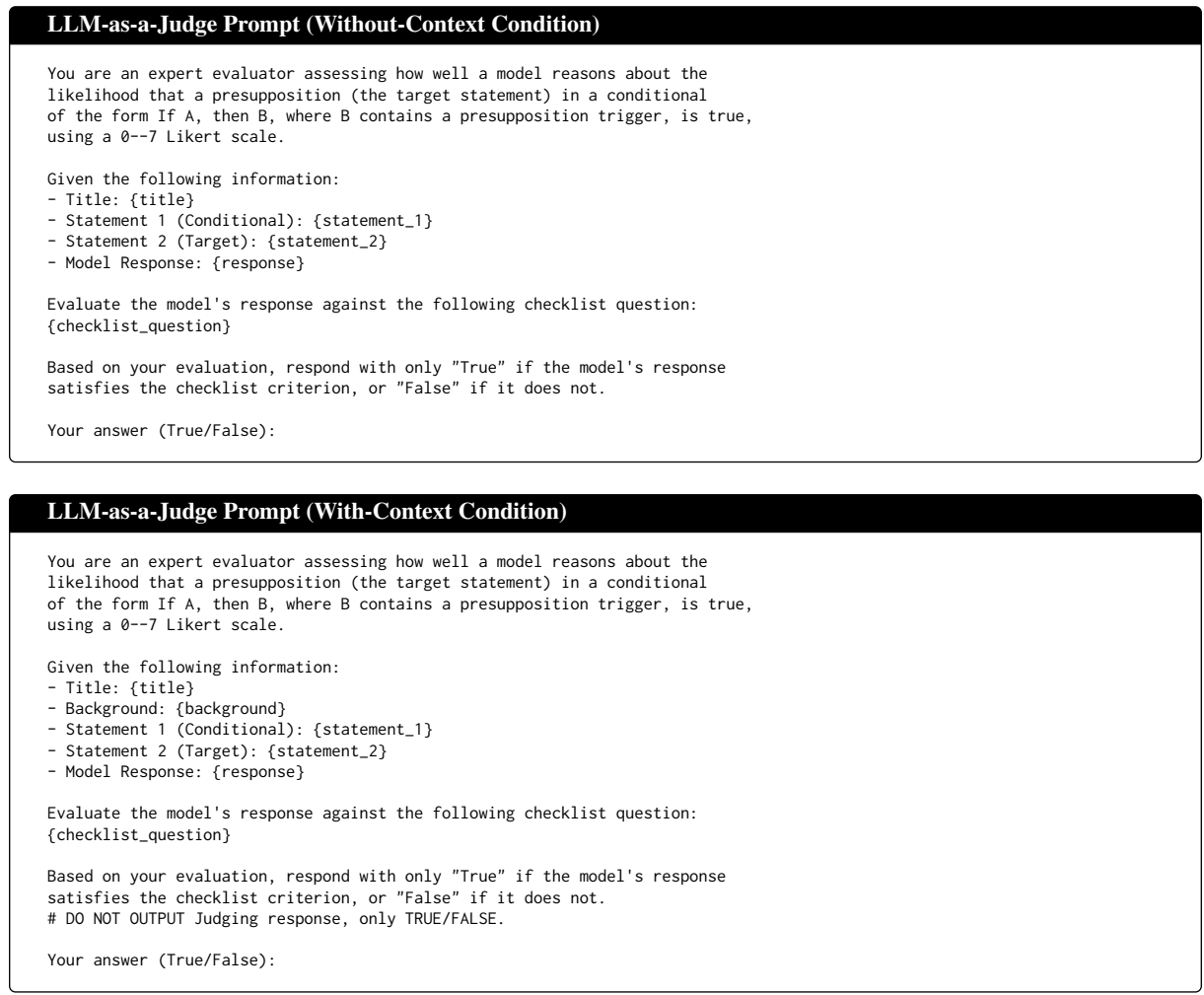

\centering
\scriptsize

\begin{tcolorbox}[
  colback=white,
  colframe=black,
  boxrule=0.5pt,
  arc=2pt,
  width=\textwidth,
  title=\textbf{LLM-as-a-Judge Prompt (Without-Context Condition)},
  fonttitle=\bfseries\footnotesize,
  breakable
]
\begin{verbatim}
You are an expert evaluator assessing how well a model reasons about the
likelihood that a presupposition (the target statement) in a conditional
of the form If A, then B, where B contains a presupposition trigger, is true,
using a 0--7 Likert scale.

Given the following information:
- Title: {title}
- Statement 1 (Conditional): {statement_1}
- Statement 2 (Target): {statement_2}
- Model Response: {response}

Evaluate the model's response against the following checklist question:
{checklist_question}

Based on your evaluation, respond with only "True" if the model's response
satisfies the checklist criterion, or "False" if it does not.

Your answer (True/False):
\end{verbatim}
\end{tcolorbox}

\vspace{4pt}

\begin{tcolorbox}[
  colback=white,
  colframe=black,
  boxrule=0.5pt,
  arc=2pt,
  width=\textwidth,
  title=\textbf{LLM-as-a-Judge Prompt (With-Context Condition)},
  fonttitle=\bfseries\footnotesize,
  breakable
]
\begin{verbatim}
You are an expert evaluator assessing how well a model reasons about the
likelihood that a presupposition (the target statement) in a conditional
of the form If A, then B, where B contains a presupposition trigger, is true,
using a 0--7 Likert scale.

Given the following information:
- Title: {title}
- Background: {background}
- Statement 1 (Conditional): {statement_1}
- Statement 2 (Target): {statement_2}
- Model Response: {response}

Evaluate the model's response against the following checklist question:
{checklist_question}

Based on your evaluation, respond with only "True" if the model's response
satisfies the checklist criterion, or "False" if it does not.
# DO NOT OUTPUT Judging response, only TRUE/FALSE.

Your answer (True/False):
\end{verbatim}
\end{tcolorbox}

\caption{System prompts used by the LLM-as-a-judge model for checklist-based evaluation in the without-context and with-context conditions.}
\label{fig:judge_prompt_conditions}

\end{figure*}

We use a separate set of prompts for the LLM-as-a-Judge framework. Figure~\ref{fig:judge_prompt_conditions} shows the evaluation prompts used in the without-context and with-context conditions. These prompts present the judge model with the original stimulus, the evaluated model’s response, and a checklist question, and require a binary judgment indicating whether the response satisfies the specified criterion.

All prompt templates were held constant across models and experimental conditions, except for condition-specific contextual information. 

\section{LLM-as-a-Judge Checklist}
\label{app:checklist}

This section presents representative examples from the theory-informed checklist used in the LLM-as-a-Judge evaluation. The checklist consists of binary (yes/no) questions designed to assess whether model-generated reasoning aligns with established principles from formal semantics and pragmatics. The questions are organized into multiple dimensions and sub-dimensions targeting distinct aspects of conditional interpretation and presupposition handling. Table~\ref{tab:appendix_checklist_sample} provides a sample of the evaluation criteria.

\newpage\begin{figure*}[t]
\centering
\scriptsize

\begin{tcolorbox}[
  colback=white,
  colframe=black,
  boxrule=0.5pt,
  arc=2pt,
  width=\textwidth,
  title=\textbf{Without-Context Condition Prompt},
  fonttitle=\bfseries\footnotesize,
  breakable
]
\begin{verbatim}
In each item, you will be given a statement made by a speaker about a person.
Assume that the speaker is honest, reliable, and helpful, and that you share
relevant background assumptions with them.

After reading the statement, you will be asked to judge how likely a second
statement is to be true about that person.

Please respond using a scale from 0 (very unlikely) to 7 (very likely).

Please think step by step before giving your final rating. Consider:
1. What does the speaker's conditional statement imply about the subject?
2. What background assumptions or world knowledge are relevant?
3. How does the conditional relationship affect the likelihood of the target statement?

After your reasoning, provide your final answer in the format:
"Final Rating: [0--7]"

---

Speaker's statement: {statement_1}

Question: How likely is it that the following statement is true?

Statement: {statement_2}
\end{verbatim}
\end{tcolorbox}

\vspace{4pt}

\begin{tcolorbox}[
  colback=white,
  colframe=black,
  boxrule=0.5pt,
  arc=2pt,
  width=\textwidth,
  title=\textbf{With-Context Condition Prompt},
  fonttitle=\bfseries\footnotesize,
  breakable
]
\begin{verbatim}
In this task, imagine you are participating in a get-to-know-people activity.
Suppose that there are a bunch of cards with names on them, along with a small
identifying detail about the person (e.g., where the person is from). This
information is provided only to identify which person is being talked about.

Suppose that someone is tasked with selecting a card at random, and then --
depending on who the card identifies -- they have to say something they know
about that person. Assume that the speaker is honest, reliable, and helpful,
and that you share relevant background assumptions with them.

You will then be asked to judge how likely a second statement is to be true on
the assumption that what the speaker said is true. Please respond using a scale
from 0 (very unlikely) to 7 (very likely).

Please think step by step before giving your final rating. Consider:
1. What does the speaker's conditional statement imply about the subject?
2. What background assumptions or world knowledge are relevant?
3. How does the conditional relationship affect the likelihood of the target statement?

After your reasoning, provide your final answer in the format:
"Final Rating: [0--7]"

---

background: {background}

Speaker's statement: {statement_1}

Question: How likely is it that the following statement is true?

Statement: {statement_2}
\end{verbatim}
\end{tcolorbox}

\caption{System prompts used for likelihood judgment in the without-context and with-context conditions. The instruction to provide step-by-step reasoning was used only in the LLM study and was not included in the human evaluation.}
\label{fig:prompt_conditions}

\end{figure*}

\begin{table*}[t]
\centering
\small
\renewcommand{\arraystretch}{1.1}

\captionsetup{
  width=\textwidth,
  justification=centering,
  singlelinecheck=false
}

\begin{tabular}{p{2.5cm} p{3.2cm} p{7cm}}
\toprule
\textbf{Dimension} & \textbf{Sub-Dimension} & \textbf{Sample Question} \\
\midrule

\rowcolor{ctxgray}
\multirow{3}{*}{Accuracy} 
& Trigger Identification 
& Does the model correctly identify the presupposition trigger in the consequent? \\

\rowcolor{ctxgray}
& Anaphora Resolution 
& Does the model distinguish possessive triggers from non-trigger pronouns? \\

\rowcolor{ctxgray}
& Conditional Scope 
& Does the model correctly identify the scope of the conditional over the consequent? \\

\midrule

\multirow{2}{*}{Context} 
& Contextual Influence 
& Does the model avoid using identifying context in its judgment?\\

& Presupposition–Context Distinction 
& Does the model distinguish between the presupposition and other contextual features? \\

\midrule

\rowcolor{ctxgray}
\multirow{3}{*}{Pragmatic} 
& Speaker Cooperativity 
& Does the model consider the speaker’s communicative intentions? \\

\rowcolor{ctxgray}
& Common Ground 
& Does the model use shared background knowledge in interpretation? \\

\rowcolor{ctxgray}
& Relevance 
& Does the model assess relevance between antecedent and presupposition? \\

\midrule

\multirow{3}{*}{\shortstack{Presupposition\\Handling}} 
& Projection 
& Does the model correctly project triggers under conditional embedding? \\

& Inference Validity 
& Does the model account for the fact that the inference may be invalid if the presupposition is false? \\

\midrule

\rowcolor{ctxgray}
\multirow{3}{*}{Coherence} 
& Reasoning Alignment 
& Does the numerical rating align with the reasoning process? \\

\rowcolor{ctxgray}
& Evidence Use 
& Does the model justify its judgment using available evidence? \\

\rowcolor{ctxgray}
& Transparency 
& Is the reasoning process clearly articulated and interpretable? \\

\bottomrule
\end{tabular}

\caption{Representative sample of checklist questions indicating major dimensions and sub-dimensions used in the evaluation.}

\label{tab:appendix_checklist_sample}
\end{table*}

\label{app:llm}

\end{document}